\pdfoutput=1

\documentclass[11pt]{article}

\usepackage[final]{acl}

\usepackage{times}
\usepackage{latexsym}
\usepackage{amssymb}
\usepackage{booktabs}

\usepackage[T1]{fontenc}

\usepackage[utf8]{inputenc}

\usepackage{microtype}

\usepackage{inconsolata}

\usepackage{adjustbox}

\usepackage{graphicx}

\usepackage{todonotes}

\newcommand{\quotes}[1]{``#1''}
\renewcommand{\vec}[1]{\mathbf{#1}}

\usepackage{fancyhdr}
\fancypagestyle{firstpage}{
  \fancyhf{} 
  \fancyfoot[C]{\thepage} 
  \fancyfoot[L]{\scriptsize Accepted at SemEval 2025. Official version: \href{https://aclanthology.org/2025.semeval-1.31/}{aclanthology.org/2025.semeval-1.31/}} 
}

\title{UWBa at SemEval-2025 Task 7: Multilingual and Crosslingual Fact-Checked Claim Retrieval}

\author{
 \textbf{Ladislav Lenc\textsuperscript{1,2}},
 \textbf{Daniel C\'{i}fka\textsuperscript{1}},
 \textbf{Ji\v{r}\'{i} Mart\'{i}nek\textsuperscript{1,2}},
 \textbf{Jakub \v{S}m\'{i}d\textsuperscript{1,2}},
\\
 \textbf{Pavel Kr\'{a}l\textsuperscript{1,2}}
\\
\\
 \textsuperscript{1}Department of Computer Science and Engineering,
  University of West Bohemia in Pilsen\\
 \textsuperscript{2}New Technologies for the Information Society, 
  University of West Bohemia in Pilsen
\\
\texttt{\{llenc,dcifka20,jimar,jaksmid,pkral\}@kiv.zcu.cz}
}

\begin{document}
\maketitle
\thispagestyle{firstpage} 
\begin{abstract}
This paper presents a zero-shot system for fact-checked claim retrieval. 
We employed several state-of-the-art large language models to obtain text embeddings.
The models were then combined to obtain the best possible result.
Our approach achieved 7th place in monolingual and 9th in cross-lingual subtasks.
We used only English translations as an input to the text embedding models since multilingual models did not achieve satisfactory results.
We identified the most relevant claims for each post by leveraging the embeddings and measuring cosine similarity.
Overall, the best results were obtained by the NVIDIA NV-Embed-v2 model.
For some languages, we benefited from model combinations (NV-Embed \& GPT or Mistral).
\end{abstract}

\section{Introduction}
The SemEval-2025 shared task 7, \textit{Multilingual and Cross-lingual Fact-Checked Claim Retrieval}~\cite{semeval2025task7}, focuses on efficiently identifying fact-checked claims across multiple languages. 
This challenge is particularly important in the fight against global misinformation \cite{khraisat2025survey, abdali2024multi}, as manual verification of claims in different languages is both time-consuming and impractical.
The task aims to support fact-checkers by developing systems that retrieve relevant, previously fact-checked claims for social media posts, addressing the complexities of a multilingual and cross-lingual context.
The task utilizes an enhanced version of the MultiClaim dataset~\cite{pikuliak2023multilingual}, specifically tailored to meet these multilingual needs. 

Our system uses a zero-shot approach based on pre-trained Text Embedding Models (TEMs). 
We selected three TEMs according to preliminary experiments and used their combination to further improve the results and make the system more robust.
All models and combinations were evaluated on the development data.
The final approach employs the best model or combination for each language.

We use all available text (incl. OCR) as input to have a maximal context. In some cases, though, we encountered model limits regarding maximum input size. The input was truncated to fit the model's tokenizer in such a case.

Based on preliminary experiments where multilingual models used with original texts achieved worse results, we use only English translations in the final approach.

Our contributions are as follows: 1) We compare several embedding models and show that simple zero-shot embeddings are effective for the task. 2) We demonstrate that larger models consistently outperform smaller ones. 3) We propose combining different models for different languages, which improves overall performance.

The following section highlights the most essential facts about the task and related work focuses on TEMs. Then, we present our approach, experimental setup, and results, including the error analysis. 
The final section concludes the paper.

\section{Background}
The basis for this SemEval-2025 task was presented in the paper \textit{Multilingual Previously Fact-Checked Claim Retrieval} \cite{pikuliak2023multilingual}. The motivation for this task is to reliably identify previously fact-checked claims for various social media posts across multiple languages.
Furthermore, this task addresses the challenge of both multilingual and cross-lingual settings, providing valuable support to fact-checkers and researchers in combating the global spread of misinformation.

The dataset for evaluation has been derived from the existing MultiClaim dataset \cite{pikuliak2023multilingual} with some specific modifications. The dataset contains tens of thousands of multilingual social media posts, matched with over 200,000 fact-checks across nearly 40 different languages.

In addition to the text of the posts, OCR output is also available (when a post includes an image), including language identification (if the image contains multilingual texts). Moreover, there are English translations for all texts and also a verdict that shows a label of the post (e.g. false or partly false information, altered photo, etc.) and a list of timestamps. 

\subsection{Related Work}
The proposed approach builds on the work of \citet{pikuliak2023multilingual} and employs various TEMs.
TEMs have become a cornerstone in natural language processing (NLP), with various approaches being developed to enhance their effectiveness and efficiency. One prominent method is the use of contrastive learning, which has been shown to improve the performance of text embeddings significantly. For instance, the General Text Embeddings (GTE) model employs multi-stage contrastive learning to unify various NLP tasks into a single format, achieving substantial performance gains over existing models by leveraging diverse datasets \cite{Li2023Towards}.

Siamese networks have also played a crucial role in the development of text embedding models. These networks are designed to learn semantically meaningful embeddings by comparing pairs of inputs. For example, the Sentence-BERT (SBERT) model utilizes a Siamese network structure to derive sentence embeddings that can be efficiently compared using cosine similarity, drastically reducing computational overhead while maintaining high accuracy \cite{Reimers2019Sentence-BERT}. 

Additionally, the Pseudo-Siamese network Mutual Learning (PSML) framework addresses the overfitting issues in contrastive learning by employing mutual learning between two encoders, thus enhancing the stability and generalization of sentence embeddings \cite{Xie2022Stable}.

Triplet loss is another technique that has been effectively integrated into text embedding models to improve their discriminative power. In the context of intention detection, a Siamese neural network with triplet loss is used to construct robust utterance feature embeddings, which are crucial for accurately identifying user intentions in dialogue systems \cite{Ren2020Intention}. 
This approach leverages metric learning to map sequence utterances into a compact Euclidean space, facilitating the distinction between similar and dissimilar inputs.

In summary, the development of text embedding models has been significantly advanced by integrating contrastive learning, Siamese networks, and triplet loss. These approaches have not only improved the performance of text embeddings across various NLP tasks but also enhanced their efficiency and applicability in real-world scenarios.

\section{The Proposed Approach}
\label{sec:system}
The system operates in a zero-shot setting, leveraging multiple TEMs to obtain sentence representations.
Specifically, we employ NVIDIA 
NV-Embed-v2 (NV-Embed)~\citep{lee2025nvembedimprovedtechniquestraining}, base multilingual GTE (mGTE)~\citep{zhang-etal-2024-mgte}, large English GTE~\citep{zhang-etal-2024-mgte}, GPT text-embedding-3-large~\citep{gptembed}, and Mistral mistral-embed~\citep{mistralembeddings}.
These models are based on the Transformer architecture~\citep{NIPS2017_3f5ee243}, which processes an input sentence $s=\{x_i\}_{i=1}^n$ of $n$ tokens and produces vectors $\vec{h} = \{\vec{h}_i\}_{i=1}^n$, where $\vec{h}_i$ is a hidden feature representation for a~corresponding token $x_i$. 

Several techniques exist to obtain a vector representation of a full sentence. 
The GTE models prepend a special \texttt{[CLS]} token at the beginning of the sequence, which serves as a global representation of the sentence. 
NV-Embed utilizes a latent attention layer to generate the final sentence-level vector.
Another common approach is a mean pooling, which averages the token-level vectors to produce a single representation. 

The NV-Embed model requires a prompt to be prefixed to input queries (in our case, posts). We use the following prompt: \textit{\quotes{Given a post, retrieve claims that verify the post}}.

For each input post, we concatenate the original text with an OCR-extracted text. We use English translations for all models except for mGTE, where we use texts in their original language. We feed the concatenated text into the models without additional preprocessing.

Similarly, we feed the fact-checks into a TEM concatenating the title and the claim. As a result, we obtain a vector $x_{post} \in \mathbb{R}^{E}$ and a matrix $\mathbb{X}_{claims}$ with a shape $n \times E$, where $n$ is the number of candidate fact-checks (claims) and $E$ is the embedding dimension based on the utilized TEM. The goal is to find the 10 closest rows (vectors) in the matrix based on the cosine similarity.

\subsection{Model Selection and Combination}
\label{sec:system_model_selection}
We select the most effective model or model combination for each language based on development data results and deploy it in the final system.
When combining two models, we select the five most similar claims for each TEM and put them together. 
If the resulting set contains duplicities, we remove them and add claims from positions six and further to build the final set of ten retrieved claims.
In the case of three models, we select only three most similar claims from each model. 
Again, we remove the possible duplicities and add claims at lower positions to get the ten required claims.
The addition of claims is done by picking one claim from the best model, then one from the second best, etc.

For the cross-lingual scenario, only the NV-embed model was used, as the combinations did not lead to improvement.

The models or model combinations used for individual languages are summarized in Table~\ref{tab:models}.

\begin{table}[htb!]
{\footnotesize
\centering
\begin{tabular}{lc}
\toprule
   \textbf{Language} & \textbf{Model/Combination}  
    \\\midrule
    ara & GPT \& NV-Embed \\
    deu & NV-Embed \\
    eng & NV-Embed \\
    fra & GPT \& NV-Embed \\
    msa & GPT \& NV-Embed \\
    pol & NV-Embed \\
    por & NV-Embed \\
    spa & GPT \& NV-Embed \\
    tha & Mistral \& NV-Embed \\
    tur & NV-Embed \\
\bottomrule
\end{tabular}
\caption{Models used for individual languages.}
\label{tab:models}
}
\end{table}

\section{Experimental Setup}

\subsection{Implementation Details}
For NV-Embed and (m)GTE, we utilize models from the HuggingFace Transformers library\footnote{\url{https://github.com/huggingface/transformers}}~\citep{wolf-etal-2020-transformers}. All experiments for these models are conducted on a single NVIDIA L40 GPU with 48 GB of memory. To accommodate NV-Embed within memory constraints, we apply 4-bit quantization.
In the case of GPT~\citep{gptembed} and Mistral~\citep{mistralembeddings}, we use the official APIs to obtain the embeddings.

\subsection{Dataset}
The dataset provided by the task organizers is an updated version of the Multiclaim dataset. During the competition, we had labels only for training data. At the end of the development phase, the organizers released the ground truth labels for development data. Since our approach is zero-shot (neither training nor fine-tuning any model on training data), we limit ourselves to only describing the development (DEV) data in this section.

\subsubsection{Development Data}
The vast majority of posts in the data are connected with only one fact-checked claim, while having three or four claims per post is rare.

The monolingual data contains eight languages, with a total of 1,891 posts. 
Table \ref{tab:dev_data_stats_2} reveals posts and pair counts for individual languages, showing a higher number of pairs for English, Portuguese, and Spanish. The opposite situation is in \textit{tha} and \textit{ara}, where the number of posts equals the number of pairs, meaning there is only a single claim relevant to the post. 

For the monolingual scenario, seeking the most relevant fact-checks is limited to the language  in which the post is written. Consequently, we expect much better results for this monolingual subtask since the set of potential candidates is much smaller, reducing the likelihood of false positives. All of our evaluations confirm this assumption, as presented below.

\begin{table}[htb!]
{\footnotesize
\centering
\begin{tabular}{lcc}
\toprule
   Language & Posts count & All pairs count
    \\\midrule
    fra & 188 & 200  \\
    spa & 615 & 692  \\
    eng & 478 & 627  \\
    por & 302 & 403  \\    
    tha & 42 & 42  \\ 
    deu & 83 & 101  \\    
    msa & 105 & 116  \\
    ara & 78 & 78  \\
\bottomrule
\end{tabular}
\caption{Development data -- number of pairs and posts for individual languages.}
\label{tab:dev_data_stats_2}
}
\end{table}

The cross-lingual dataset consists of 552 posts and 651 pairs, with no language-specific information since the task is to find relevant claims across all languages.

\section{Experiments}

The task organizers have chosen success-at-K (S@K) as a main evaluation metric. The metric expresses a percentage of pairs when at least one of the desired fact-checks ends up in the top K, where K = 10 for this task. The S@K for each particular language is computed separately, and the mean value represents the multilingual result. 

\subsection{Results}

According to the official test leaderboard\footnote{\url{https://www.codabench.org/competitions/3737/\#/pages-tab}}, our approach achieved 7th place for the monolingual and 9th for the cross-lingual scenario, respectively. There are 28 participants in the monolingual subtask and 29 in the cross-lingual subtask.
Table~\ref{tab:test_data_all_avg_results} presents the results of our system on the test data.

\begin{table}[htb!]
\centering
\begin{adjustbox}{width=\linewidth}
\begin{tabular}{lccccc}
\toprule
   \textbf{Scenario} & \textbf{Comb.} & \textbf{NV-Embed} & \textbf{GPT} & \textbf{Mistral} & \textbf{Best}
    \\\midrule
    Monoling. & \textbf{0.927} (7.) & 0.919 & 0.903 & 0.902 & \textit{0.960} \\
    Cross-ling. & - & \textbf{0.783} (9.) & 0.741 & 0.745 & \textit{0.859}  \\
\bottomrule
\end{tabular}
\end{adjustbox}
\caption{Average S@10 monolingual and cross-lingual results on test data, our best result are in \textbf{bold}, its rank in parentheses, the \textbf{Best} column shows the highest score achieved in the competition. The \textbf{Comb.} column represents the model combination described in Section \ref{sec:system_model_selection}.}

\label{tab:test_data_all_avg_results}
\end{table}

\begin{table}[htb!]
{\footnotesize
\centering
\begin{tabular}{lccc}
\toprule
   Model & \textbf{S@10} & \textbf{S@5} & Dif
    \\\midrule
    NV-Embed & 0.775 & 0.672 & -13.29\% \\
    GPT & 0.726 & 0.627 & -13.64\% \\
    Mistral & 0.719 & 0.612 & -14.88\% \\
\bottomrule
\end{tabular}
\caption{Comparison of S@5 and S@10 results for cross-lingual subtask on development data.}
\label{tab:dev_data_results}
}
\end{table}

The S@10 metric puts the most relevant fact-checks for a given post, which ended up in the top 3, for example, and those that barely fit into the first 10, on the same level. Once the DEV data were released, we computed S@5 for our models to investigate the behavior of the models when the ``harder'' metric is used. The greater the drop in this metric compared to S@10, the less likely the correct pair will be among the most relevant results. Conversely, if the decrease is minimal, it indicates that the models are performing well, placing the most relevant fact-checks in the top positions. Tables \ref{tab:dev_data_results} and \ref{tab:dev_data_monoling_results} show such a comparison; the decrease is evident for all models. 

We have an interesting observation in monolingual results. Even though we used English translations, the percentage difference between S@10 and S@5 varies significantly for the individual languages (compare, for example, \textit{eng} or \textit{por} with \textit{fra} in Table \ref{tab:dev_data_monoling_results} in the Appendix).

\subsection{Model Comparison}
Table~\ref{tab:dev_data_all_avg_results} presents the average monolingual and cross-lingual S@10 results on the development data. The GTE and mGTE models performed significantly worse than the other models, particularly in cross-lingual settings, so we excluded them from further experiments. For mGTE, we used the original language data, as it is a multilingual model. 

For all other models, which are primarily English-centric, we used English-translated data. We attribute the poor performance of the GTE models to their smaller parameter sizes (approximately 350M for mGTE and 434M for GTE) compared to the other models. However, their smaller size and open-access nature make them easier to deploy, with higher inference speeds. In contrast, NV-Embed has about 7B parameters, while GPT and Mistral require API access.

\begin{table}[htb!]
\centering
\begin{adjustbox}{width=\linewidth}
\begin{tabular}{lccccc}
\toprule
   \textbf{Scenario} & \textbf{GTE} & \textbf{mGTE} & \textbf{NV-Embed} & \textbf{GPT} & \textbf{Mistral}
    \\\midrule
    Monolingual & 0.777 & 0.785 & \textbf{0.902} & 0.856 & 0.863 \\
    Cross-lingual & 0.569 & 0.574 & \textbf{0.775} & 0.726 & 0.719 \\
\bottomrule
\end{tabular}
\end{adjustbox}
\caption{Average S@10 scores of TEM models on development data. All texts are translated to English except for mGTE, which uses the original language. Best results are shown in \textbf{bold}.}
\label{tab:dev_data_all_avg_results}
\end{table}

Among the remaining models, NV-Embed achieved the highest average performance, with Mistral and GPT following closely.

Table~\ref{tab:dev_data_combinations} shows the performance of various model combinations for individual languages on the development set. These results guided the selection of the best-performing combinations for our final system.

\begin{table*}[ht!]
{\footnotesize
\centering
\begin{tabular}{@{}lccccccccc@{}}
\toprule
Model/Combination & eng & fra & deu & por & spa & tha & msa & ara & avg \\
\midrule
GPT & 0.85 & 0.92 & 0.70 & 0.83 & 0.89 & 0.98 & 0.88 & 0.82 & 0.86 \\
Mistral & 0.84 & 0.90 & 0.80 & 0.82 & 0.90 & 0.98 & 0.88 & 0.81 & 0.86 \\
NV-Embed & \textbf{0.87} & 0.95 & \textbf{0.89} & \textbf{0.88} & 0.92 & 0.95 & 0.90 & 0.86 & 0.90 \\
GPT \& Mistral & 0.83 & 0.90 & 0.75 & 0.81 & 0.89 & 0.95 & 0.88 & 0.79 & 0.85 \\
GPT \& NV-Embed & 0.87 & \textbf{0.95} & 0.88 & 0.87 & \textbf{0.92} & 0.95 & \textbf{0.90} & \textbf{0.87} & 0.90\\ 
Mistral \& NV-Embed & 0.87 & 0.95 & 0.88 & 0.87 & 0.92 & \textbf{0.98} & 0.89 & 0.86 & 0.90 \\
GPT \& Mistral \& NV-Embed & 0.87 & 0.93 & 0.84 & 0.86 & 0.92 & 0.98 & 0.90 & 0.86 & 0.89 \\
\bottomrule
\end{tabular}
\caption{Results of model combinations on the development data.}
\label{tab:dev_data_combinations}
}
\end{table*}

\subsection{Error Analysis}
\label{sec:error_analysis}
Since our final approach utilizes solely English translations, the error analysis only focuses on the cross-lingual scenario where a candidate fact-checks space is not limited to a particular language. 

\begin{figure}[htb!]
    \centering
    \includegraphics[width=\linewidth]{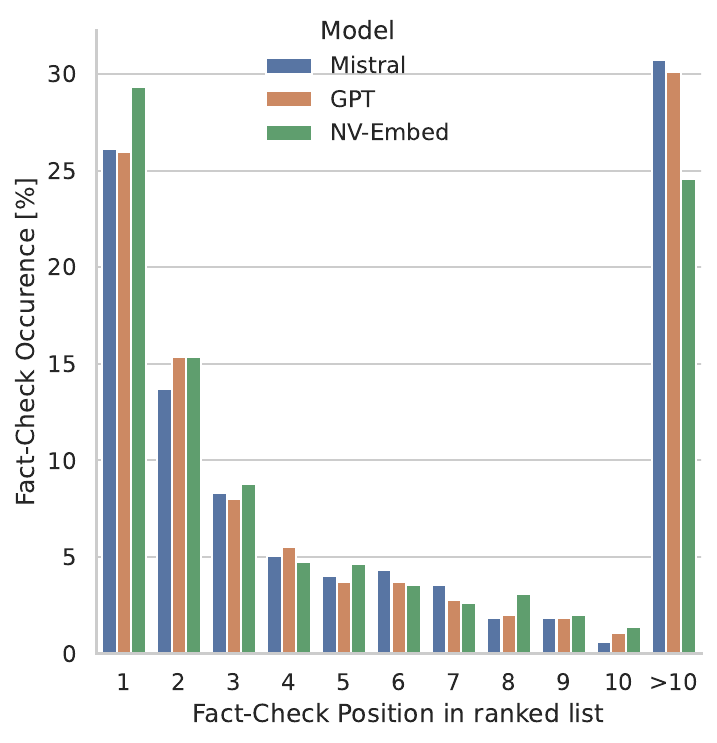}
    \caption{Position of the searched fact-check among the most similar fact-checks for cross-lingual subtask.}    \label{fig:gpt_mistral_nv_embed_analysis}
\end{figure}

As illustrated in Figure \ref{fig:gpt_mistral_nv_embed_analysis}, all three embedding models correctly assign over a quarter of fact-checks to the top 1 ranked position. Furthermore, in the top 3 results, each model accurately retrieves around 50\% of fact-checks. In other words, when a model identifies the correct ``post-to-fact-check'' pair, it typically ranks it within the top 3 positions, demonstrating high confidence in its predictions. The number of missed fact-checks (the position in the ranked list of 11 or more) ranges from 25 to 30\% for all models.

Table \ref{tab:fact_checks_missed} presents the number of missed fact-checks per post. This metric is comparable to the official S@10 score, with the key difference being its focus on individual pairs. The numbers in the table reflect all correctly assigned fact-checks for each post, not just whether at least one was correctly matched.

\begin{table}[htb!]
{\footnotesize
\centering
\begin{tabular}{lcc}
\toprule
& \textbf{Missed} & \textbf{Fact-checks}\\
   Model & \textbf{Fact-checks} & \textbf{in Top 10}  
    \\\midrule
    NV-Embed & 160  (24.6\%) & 491  (75.4\%)\\
    GPT & 196  (30.2\%) & 455  (69.8\%)\\
    Mistral & 200  (30.7\%) & 451  (69.3\%)\\
\bottomrule
\end{tabular}
\caption{Missed pairs and true positive pairs within a top-10 selection ranked list.}
\label{tab:fact_checks_missed}
}
\end{table}

\section{Conclusion}
This paper described our approach for the SemEval-2025 shared task 7 \textit{Multilingual and Cross-lingual Fact-Checked Claim Retrieval}.
We adopted a zero-shot approach using large language models like NVIDIA NV-Embed-v2, GPT text-embedding-3-large, and Mistral.
This approach allows seamless integration of new languages without retraining, as demonstrated when Polish and Turkish were added to the test set.
By leveraging the embeddings and measuring cosine similarity, we identified the most relevant claims for each post.

Our approach ranked 7th out of 28 in the monolingual subtask and 9th out of 29 in the cross-lingual subtask.

Error analysis showed that all three models effectively placed the most relevant claims at the top of the ranked lists. For some languages, combining models improved performance, and our final submission reflected this. Among the models, NV-Embed proved the most effective, keeping the number of missed pairs below 25\% in the cross-lingual setting.

\section*{Acknowledgments}
The work of the students, Daniel Cífka and Jakub Šmíd, was supported by the Grant no. SGS-2025-022 - New Data Processing Methods in Current Areas of Computer Science.
The work of the other authors was supported by the project R\&D of Technologies for Advanced Digitalization in the Pilsen Metropolitan Area (DigiTech) No. CZ.02.01.01/00/23\_021/0008436. 
\bibliography{custom}

\appendix

\section{Appendix}
\label{sec:appendix}

\begin{figure}[htb!]
    \centering
    \includegraphics[width=\linewidth]{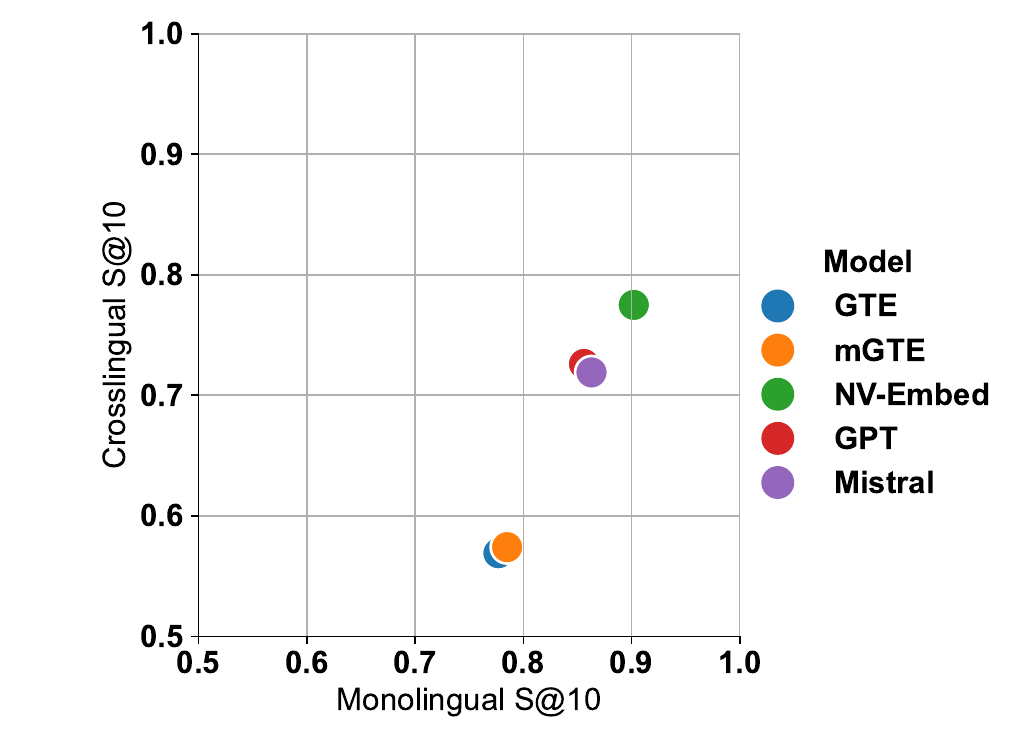}
    \caption{Comparison of monolingual and cross-lingual results of TEM models on development data. Monolingual S@10 labels the average S@10 for all languages.}
    \label{fig:model_comparison}
\end{figure}

\begin{figure}[htb!]
    \centering
    \includegraphics[width=\linewidth]{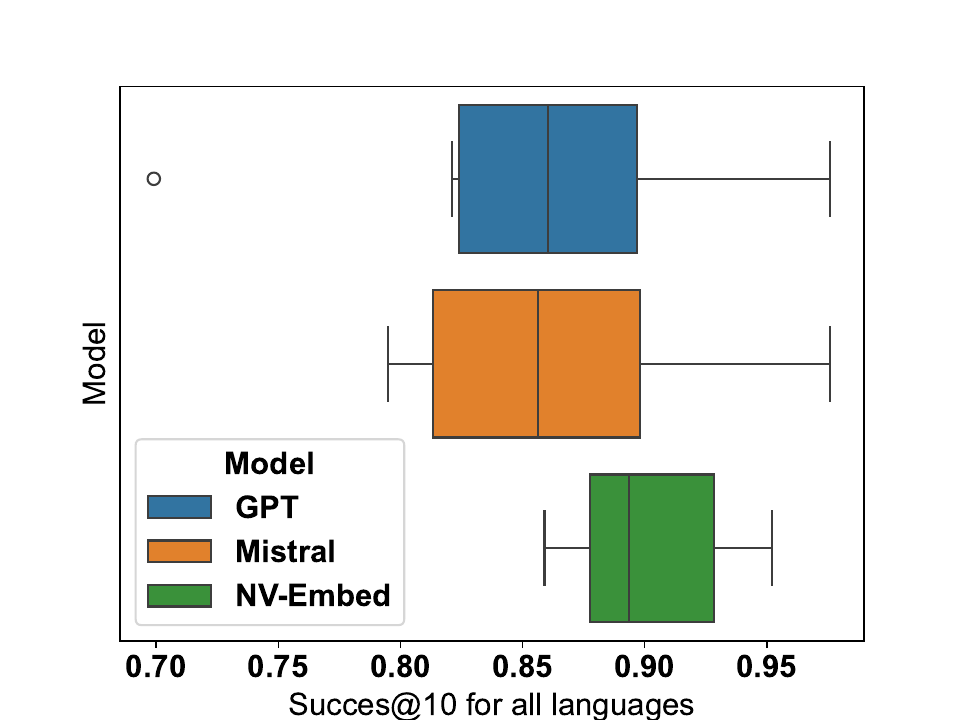}
    \caption{Development data monolingual comparison. NV-Embedd has a smaller variance and is more consistent across languages.}
    \label{fig:model_comparison_2}
\end{figure}

This appendix continues the error analysis and shows additional figures and tables.
Figure \ref{fig:model_comparison} depicts the comparison of monolingual and cross-lingual results showing the dominance of the NV-Embed model.
Monolingual results are depicted in Figure \ref{fig:model_comparison_2}. The boxplot shows average S@10 results. The width of the individual boxes reflects the monolingual results and variance with a median value represented by the vertical line inside the box. The figure confirms the supremacy of the NV-Embed model over the others.

Lastly, we present detailed development data results in Tables \ref{tab:dev_data_monoling_results} and \ref{tab:dev_data_monoling_results_pairs_evaluation}. We recalculated the S@10 and S@5 while the number of pairs was considered (Table \ref{tab:dev_data_monoling_results_pairs_evaluation}) to show a higher difficulty of this task. In this setting, all fact-checks must be in the ranked list (top 10) to be considered as a correct sample. Naturally, the numbers, in general, are lower than the official results. Since \textit{eng} and \textit{por} have more fact-checks than other languages, the discrepancy between S@10 and S@5 is much bigger.

\begin{table*}[ht!]
{\footnotesize
\centering
\begin{tabular}{@{}lccccccccc@{}}
\toprule
     & \multicolumn{3}{c}{\textbf{GPT}}          & \multicolumn{3}{c}{\textbf{Mistral}}      & \multicolumn{3}{c}{\textbf{NV-Embed}}     \\ \cline{2-10} 
Lang & S@10           & S@5            & Dif     & S@10           & S@5            & Dif     & S@10           & S@5            & Dif     \\ \hline
fra  & 0.915          & 0.888          & -2.95\% & 0.899          & 0.872          & -3.00\% & \textbf{0.947} & \textbf{0.931} & -1.69\% \\
spa  & 0.891          & 0.854          & -4.15\% & 0.898          & 0.852          & -5.12\% & \textbf{0.922} & \textbf{0.878} & -4.77\% \\
eng  & 0.845          & 0.789          & -6.63\% & 0.837          & 0.768          & -8.24\% & \textbf{0.868} & \textbf{0.801} & -7.72\% \\
por  & 0.825          & 0.765          & -7.72\% & 0.815          & 0.781          & -4.17\% & \textbf{0.881} & \textbf{0.821} & -6.81\% \\
tha  & \textbf{0.976} & \textbf{0.952} & -2.46\% & \textbf{0.976} & \textbf{0.952} & -2.46\% & 0.952          & 0.929          & -2.42\% \\
deu  & 0.699          & 0.675          & -3.43\% & 0.795          & 0.783          & -1.51\% & \textbf{0.892} & \textbf{0.843} & -5.49\% \\
msa  & 0.876          & 0.838          & -4.34\% & 0.876          & \textbf{0.857} & -2.17\% & \textbf{0.895} & \textbf{0.848} & -5.25\% \\
ara  & 0.821          & 0.769          & -6.33\% & 0.808          & 0.756          & -6.44\% & \textbf{0.859} & \textbf{0.808} & -5.94\% \\ \hline
avg  & 0.856          & 0.816          & -4.64\% & 0.863          & 0.828          & -4.06\% & \textbf{0.902} & \textbf{0.858} & -4.99\% \\ \bottomrule
\end{tabular}     
\caption{S@5 and S@10 monolingual results on development data together with a percentage difference. Best results for each language are in \textbf{bold}.}
\label{tab:dev_data_monoling_results}
}
\end{table*}

\begin{table*}[ht!]
{\footnotesize
\centering
\begin{tabular}{@{}lccccccccc@{}}
\toprule
     & \multicolumn{3}{c}{\textbf{GPT}} & \multicolumn{3}{c}{\textbf{Mistral}} & \multicolumn{3}{c}{\textbf{NV-Embed}} \\ \cmidrule(lr){2-4} \cmidrule(lr){5-7} \cmidrule(lr){8-10}
Lang & S@10     & S@5      &  Dif          & S@10      & S@5       &  Dif            & S@10       & S@5       &  Dif            \\ \midrule
fra  & 0.915    & 0.890     & -2.73\%    & 0.900     & 0.875     & -2.78\%      & \textbf{0.945}      & \textbf{0.930}     & -1.59\%      \\
spa  & 0.877    & 0.832    & -5.13\%    & 0.884     & 0.840     & -4.98\%      & \textbf{0.910}      & \textbf{0.866}     & -4.84\%      \\
eng  & 0.802    & 0.708    & -11.72\%   & 0.794     & 0.694     & -12.59\%     & \textbf{0.833}      & \textbf{0.740}     & -11.16\%     \\
por  & 0.794    & 0.720    & -9.32\%    & 0.799     & 0.730     & -8.64\%      & \textbf{0.854}      & \textbf{0.772}     & -9.60\%      \\
tha  & \textbf{0.976}    & \textbf{0.952}    & -2.46\%    & \textbf{0.976}     & \textbf{0.952}     & -2.46\%      & 0.952      & 0.929     & -2.42\%      \\
deu  & 0.703    & 0.681    & -3.13\%    & 0.782     & 0.752     & -3.84\%      & \textbf{0.881}      & \textbf{0.832}     & -5.56\%      \\
msa  & 0.862    & 0.810    & -5.81\%    & 0.853     & \textbf{0.828}     & -2.93\%      & \textbf{0.887}      & \textbf{0.828}     & -6.65\%      \\
ara  & 0.821    & 0.769    & -6.33\%    & 0.808     & 0.756     & -6.44\%      & \textbf{0.859}      & \textbf{0.808}     & -5.94\%      \\ \hline
avg  & 0.844    & 0.796    & -5.69\%    & 0.850     & 0.803     & -5.53\%      & \textbf{0.890}      & \textbf{0.838}     & -5.84\%      \\ \bottomrule
\end{tabular}
\caption{Development data S@5 and S@10 monolingual comparison when pairs are considered (all fact-checks must be ranked in top k to be considered as a correct sample). The best S@5 and S@10 results for each language are in \textbf{bold}.}
\label{tab:dev_data_monoling_results_pairs_evaluation}
}
\end{table*}

\end{document}